\documentclass[conference]{ieeeconf}

\IEEEoverridecommandlockouts                              

\usepackage{caption}
\usepackage{subcaption}

\usepackage{graphics} 
\usepackage{epsfig} 
\usepackage{mathptmx} 
\usepackage{times} 
\usepackage{amsmath} 
\usepackage{amssymb}  
\usepackage{bm}
\usepackage{color}
\usepackage{colortbl} 
\usepackage{array}
\usepackage{threeparttable}
\usepackage{booktabs}
\usepackage{algorithmic}
\usepackage{algorithm}
\usepackage{xcolor}
\usepackage{hyperref}
\usepackage{csquotes}

\DeclareMathOperator*{\argmin}{arg\,min}
\newtheorem{simplification}{Simplification}
\definecolor{Gray}{gray}{0.9}
\newcommand{\rev}[1]{{\textcolor{red}{#1}}}

\title{\LARGE \bf
Catch the Ball: Accurate High-Speed Motions \\for Mobile Manipulators via Inverse Dynamics Learning 
}

\author{Ke Dong$^{1}$, Karime Pereida$^{1}$, Florian Shkurti$^{2}$ and Angela P. Schoellig$^{1}$
\thanks{
$^{1}$ Ke Dong, Karime Pereida and Angela P. Schoellig are with the Dynamic Systems Lab (www.dynsyslab.org)
at the University of Toronto Institute for Aerospace Studies (UTIAS), and the Vector Institute for artificial Intelligence, Canada. Email: ke.dong, karime.pereida@robotics.utias.utoronto.ca, schoellig@utias.utoronto.ca
}
\thanks{$^{2}$ Florian Shkurti is with Department of Computer Science at the University of Toronto, Canada. Email: florian@cs.toronto.edu}
}%

\begin{document}

\maketitle
\thispagestyle{empty}
\pagestyle{empty}

\begin{abstract}
 Mobile manipulators consist of a mobile platform equipped with one or more robot arms and are of interest for a wide array of challenging tasks because of their extended workspace and dexterity. Typically, mobile manipulators are deployed in slow-motion collaborative robot scenarios. In this paper, we consider scenarios where accurate high-speed motions are required. We introduce a framework for this regime of tasks including two main components:  \emph{(i)} a bi-level motion optimization algorithm for real-time trajectory generation, which relies on Sequential Quadratic Programming (SQP) and Quadratic Programming (QP), respectively; and \emph{(ii)} a learning-based controller optimized for precise tracking of high-speed motions via a learned inverse dynamics model. We evaluate our framework with a mobile manipulator platform through numerous high-speed ball catching experiments, where we show a success rate of $\bf{85.33\%}$. To the best of our knowledge, this success rate exceeds the reported performance of existing related systems~\cite{bauml2010kinematically, bauml2011catching} and sets a new state of the art.
\end{abstract}

\section{Introduction}
\label{sec:intro}
Mobile manipulators combine mobile platforms and robot arms. Unlike fixed-base industrial arms, mobile manipulators have a larger workspace and can deal with more flexible and complex tasks such as housekeeping, construction, and airplane assembly~\cite{bodily2017motion}. Current research on mobile manipulators mainly focuses on techniques in slow-motion scenarios, such as kinematics redundancy resolution~\cite{de2006kinematic}, optimal path planning~\cite{bodily2017motion} and control~\cite{ellekilde2009control}. However, mobile manipulators are capable of high-speed motions, and thus have the potential to be used in applications such as industrial high-speed pick-and-place, or other challenging tasks such as table tennis~\cite{kocc2018online}. In this paper we focus on high-speed trajectory generation and accurate tracking for mobile manipulators, with an emphasis on not compromising accuracy at higher speeds.

We focus on a benchmark task for accurate high-speed motions: object catching, specifically, ball catching. Within a fraction of a second, the robot needs to \emph{(i)} predict the ball's trajectory, \emph{(ii)} generate and update a trajectory for graceful catching, and \emph{(iii)} accurately track it with the mobile manipulator. There are numerous papers~\cite{bauml2010kinematically, frese2001off, mirrazavi2016dynamical, lampariello2011trajectory} that address the ball catching problem with a \textit{fixed-base} manipulator. However, using a \textit{mobile} manipulator introduces additional challenges, due to the interplay between the following three characteristics:

\textbf{Controlling wheel-driven bases}: The mobile bases that are driven by Mecanum wheels~\cite{floresmodelling}, as used in this paper, usually have lower position accuracy than manipulators. One example is the position accuracy of the Kuka KMR Quantec driven by Mecanum wheels of $\pm5~mm$~\cite{kuka_base} while the accuracy of the UR10 arm is $\pm0.05~mm$~\cite{ur10_arm}. Thus extra efforts are needed to accurately control the base.

\textbf{High precision in position and time}: Precise end effector trajectory tracking, both spatially and temporally and in position and orientation, is a pre-requisite for successful catching. A precision of $16~mm$ in position and $3~ms$ in time must be achieved in this paper. These parameters vary depending on the gripper~\cite{bauml2011catching}.

\textbf{Real-time computation}: On average, the duration of the ball's free flight trajectory is one second, which puts a hard constraint on the computation time of ball prediction, motion planning and control. Motion planning is very challenging since complex constraints such as the nonlinear robot dynamics and forward kinematics must be considered.

\begin{figure}[tbp]
\centering
\includegraphics[width=\linewidth]{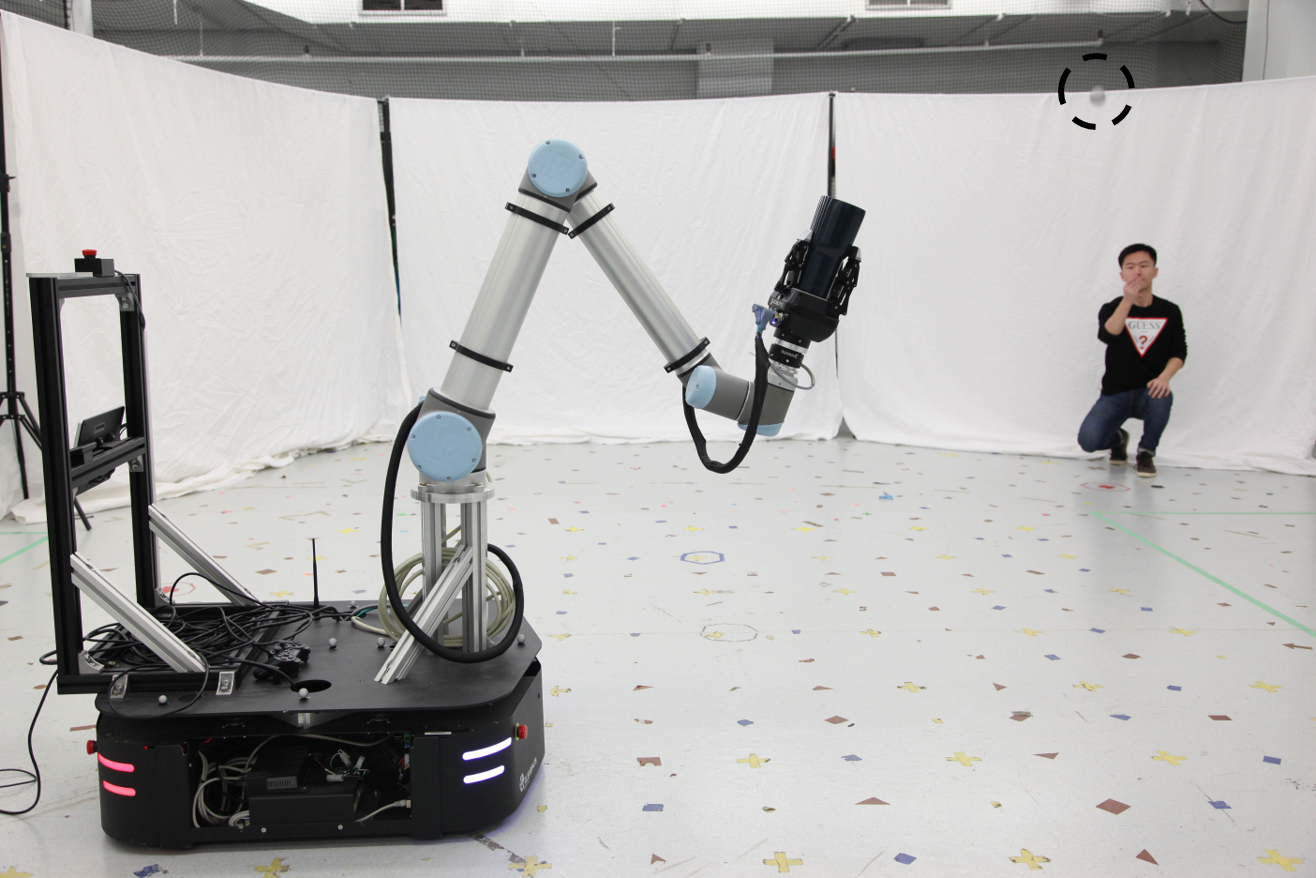}
\caption { Ball catching experimental setup. On average, the ball is thrown by a human operator from a distance of about $4~m$ towards the robot with a speed of typically $6~m/s$, resulting in a flight time of about $1~s$. A video of the experiment results is available at \url{http://tiny.cc/ball_catch}.}
\vspace{-10pt}
\label{fig:geo-setup}
\end{figure}

Given the three challenges above, accurate high-speed motion generation for mobile manipulator ball catching has not been widely explored, with a few very notable exceptions~\cite{bauml2011catching}. In this paper we introduce: \emph{(i)} a bi-level motion planning scheme that generates smooth trajectories despite nonlinear constraints in real time, and \emph{(ii)} a learning-based controller improving tracking accuracy and enabling the mobile manipulator to follow trajectories sufficiently accurately. Our proposed bi-level optimization scheme for motion planning uses sequential quadratic programming (SQP) to select the catching pose, and solves a low-level quadratic program (QP) that generates the trajectory to the catching pose. Our proposed learning-based controller learns the inverse dynamics of the system~\cite{zhou-cdc17} to improve the control accuracy of the mobile manipulator. We use a modified Kalman Filter for improved ball state estimation, which incorporates the influence of aerodynamic drag in the state prediction step. 

We present a thorough analysis and evaluation of major components of the system. Experiment data shows that our bi-level optimization scheme has an average run time of $5.1~ms$, and the inverse dynamics learning method can reduce tracking error up to $79.1\%$. A success rate of $85.33\%$ of real robot ball catching task is achieved, which is higher than the $80\%$ success rate reported in~\cite{bauml2010kinematically, bauml2011catching}. A video with experiments showing accurate high-speed motions can be found at \url{http://tiny.cc/ball_catch}.

The contributions of this paper are two-fold. First, we propose an algorithmic framework to generate and accurately follow high-speed trajectories with a mobile manipulator. The proposed framework has a bi-level optimization scheme for motion planning, which can generate smooth joint trajectories under complex constraints in real-time. Additionally, the proposed framework has a learning-based controller to improve the tracking performance of the mobile manipulator. Our second contribution is the extensive evaluation of the performance of the proposed framework on the mobile manipulator, which includes a thorough error analysis of major components. This framework is tested on real robot catching experiments and achieves a success rate of $85.33\%$. This success requires a tight interplay between different system components and hardware constraints. Implementation details of such a complex system are presented, and should a resource for similar system designs.

\section{Related Work}
\label{sec:related_work}
A large body of work has been devoted to ball catching with a fixed-base manipulator~\cite{bauml2010kinematically, frese2001off, mirrazavi2016dynamical, lampariello2011trajectory}, with special focus on motion planning. Early pioneering work in this field used heuristic methods for selecting the catching point and generated the joint trajectory via interpolation~\cite{frese2001off}. Such methods are simple, as they restrict catching points to a certain area, but may not find the optimal robot motion for catching.  Recent work uses nonlinear optimization methods for selecting a catching pose, and  parameterization methods, like trapezoidal velocity profile and third-order polynomials, for path generation for simplicity. In~\cite{bauml2010kinematically}, a ball catching system consisting of a fixed-base $7$ Degree-of-Freedom (DoF) arm and a four-finger hand is presented. It uses SQP for high-level goal selection, and restricts joint trajectories to trapezoidal velocity profiles. In contrast, in our framework, we do not parameterize the trajectory. Instead, we use a QP to generate intermediate waypoints which define smooth and flexible trajectories. In the realm of high-speed manipulation, the work in~\cite{ mirrazavi2016dynamical} proposes a control law based on dynamical systems that enables the system to catch fast-moving objects. However, only fixed-base manipulators are used.

The work most similar to ours is the mobile humanoid system presented in~\cite{bauml2011catching}, which consists of two light-weight $7$ DoF arms and an omnidirectional mobile base. In their work, the mobile base is restricted to motion in the $x$ axis only. Further, they use customized hardware capable of high-speed motions. In contrast, in our work, we achieve high-speed motions in the $x-y$ plane with the mobile base, which brings the motion capacity of omnidirectional platforms into full play. Moreover, we use an off-the-shelf mobile manipulator and a learning-based controller for improved tracking accuracy, which makes our approach easy to generalize to other robotic platforms without the need for customized hardware and low-level access.

There are numerous works that try to improve the tracking performance of high-speed and aggressive trajectories. Non-learning-based methods~\cite{ellekilde2009control,seraji1998unified} rely on accurate analytical system dynamics models, which are hard to obtain. Learning-based methods mainly focus on learning the system's dynamics or inverse dynamics models~\cite{punjani2015deep, frye2014direct}. In~\cite{frye2014direct}, a Deep Neural Network (DNN) is used for direct inverse control of a quadrotor. However, guaranteeing stability with a DNN in the control loop can be very difficult. In contrast, the inverse dynamics learning method from~\cite{zhou-cdc17} uses a DNN as an add-on module outside the original closed-loop system, which is safer in terms of stability.. This method's ability for improved, impromptu trajectory tracking has been demonstrated on quadrotors~\cite{zhou-cdc17, li2017deep}, but not on mobile manipulators.


\section{System overview and problem statement}
\label{sec:system_overview}
This section presents a system overview of the mobile manipulator and the flying ball. Finally, a formal statement of the ball catching problem is given.

\subsection{System Overview}
\begin{figure}[htb]
\centering
\includegraphics[width=0.90\linewidth]{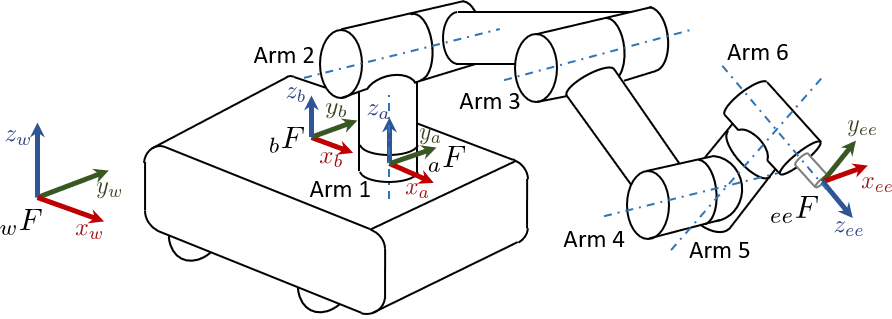}
\caption{The mobile manipulator with the world frame, $_wF$, base frame $_bF$, arm frame $_aF$ and end effector frame $_{ee}F$ is shown. Rotation axes of the six arm joints are also presented. Adapted from~\cite{Michael2017Thesis}.}
\label{fig:robot-kinematics}
\end{figure}

The mobile manipulator system considered in this paper consists of a 6-DoF manipulator rigidly mounted on a 3-DoF omnidirectional mobile base. The yaw rotation of the mobile base is not used in this work since it is too slow for the application at hand. Thus the system has 8 DoF in total. We refer to these DoF as robot joints, for brevity. The joint configuration vector of the robot is $\mathbf{q} = [\mathbf{q}_a^T, \mathbf{q}_b^T]^{\top}$, where $\mathbf{q}_a = [\theta_1, \theta_2, ...,  \theta_6]^{\top}$ is a vector containing the arm joint angles and $\mathbf{q}_b = [x_b, y_b]^{\top}$ represents the Cartesian position of the mobile base. The forward kinematics equations of the mobile manipulator shown in Fig.~\ref{fig:robot-kinematics} are assumed to be known~\cite{hawkins2013analytic}.

The flying ball's motion is modeled as a point mass under the influence of aerodynamic drag and gravity. The aerodynamic drag is proportional to the square of the speed~\cite{muller2011quadrocopter}. The ball's equation of motion can be written as follows:
\begin{equation}
    \label{eqn:ball_motion}
    \ddot{\mathbf{b}} = - \mathbf{g} - K_D||\dot{\mathbf{b}}||\dot{\mathbf{b}},
\end{equation}
where  $\mathbf{b} = [b_x, b_y, b_z]^{\top}$ is the ball's Cartesian position in the world frame and $\dot{\mathbf{b}}$, $\ddot{\mathbf{b}}$ are the corresponding velocity and acceleration, $\mathbf{g} = [0, 0, g]^{\top}$ ($g = 9.81$) is the gravitational acceleration, $K_D$ is the aerodynamic drag coefficient, and $||\cdot||$ is the Euclidean norm.

\subsection{Problem Statement}
 Given initial robot configuration $\mathbf{q}_0$, ball state $[\mathbf{b}_0^T, \dot{\mathbf{b}}_0^T]^T$, and time $t_0$, the goal is to catch the ball with the robot's end effector at a certain catch time $t_f>t_0$. To achieve this, two conditions need to be satisfied: 1) at catch time $t_f$, the origin of the end effector frame should coincide with the ball’s position; and 2) the $z$ axis of the end effector frame should be aligned with the ball’s velocity vector. Formally, these constraints can be written as:
\begin{equation}
\begin{aligned}
    \label{eqn:probblem_formulation}
    _w\mathbf{P}_{ee}(\mathbf{q}_f) &= \mathbf{b}(t_f) \\
    cos(<_w\mathbf{z}_{ee}(\mathbf{q}_f), &- \dot{\mathbf{b}}(t_f)>) = 1,
\end{aligned}
\end{equation}
where $\mathbf{q}_f$ is the robot joint configuration at catch time $t_f$, $_w\mathbf{P}_{ee}$ is the position of the end effector frame's origin in the world frame, $_w\mathbf{z}_{ee}$ is the vector of the $z$ axis of the end effector in the world frame, and $<\cdot, \cdot>$ refers to the angle between two vectors. 

\section{Methodology}
\label{sec:methodology}
\begin{figure}[!btph]
\centering
\includegraphics[width=1.0\linewidth]{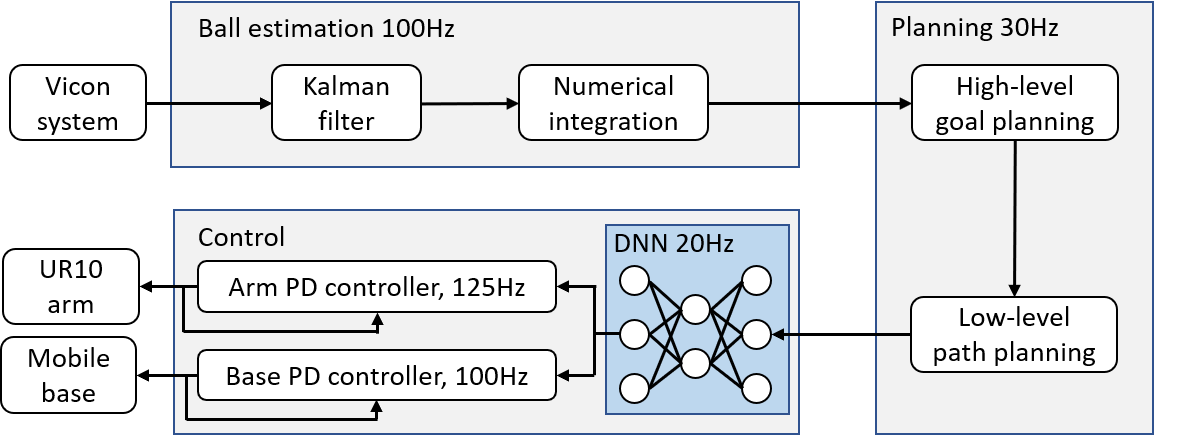}
\caption{Diagram of the overall framework for accurate high-speed ball catching. Note that the estimation module shown here is for ball state estimation, with separate, but not shown estimation modules for arm and base controllers.}
\vspace{-10pt}
\label{fig:sys-diagram}
\end{figure}

This section presents the framework we propose, which includes three main components: \emph{(i)} ball estimation and prediction, \emph{(ii)} bi-level motion planning, and \emph{(iii)} low-level control with inverse dynamics learning, as shown in Fig.~\ref{fig:sys-diagram}.

\subsection{Ball Estimation and Prediction}
\label{sec:ball_estimation_prediction}
The first component estimates the ball state and uses it to generate trajectory predictions within a finite horizon. 
To estimate the ball state $\mathbf{s}_b = [\mathbf{b}^T, \dot{\mathbf{b}}^T]^T$, we use a modified discrete-time Kalman filter that considers the influence of air drag in the state prediction step, as presented in~\cite{muller2011quadrocopter}. In the state prediction step at the discrete time index $k$, we predict the ball state $\mathbf{s}_b[k+1|k]$ at time step $k+1$ given measurements up to time $k$ according to the estimated state at the previous time step $\mathbf{s}_b[k|k]$ and motion model in \eqref{eqn:ball_motion} as follows:
\begin{equation}
\label{eqn:kalman_filter}
\begin{aligned}
    \ddot{\mathbf{b}}[k|k] &= - \mathbf{g} - K_D||\dot{\mathbf{b}}[k|k]||\dot{\mathbf{b}}[k|k] \\
    \dot{\mathbf{b}}[k+1|k] &= \dot{\mathbf{b}}[k|k] + \delta_k \ddot{\mathbf{b}}[k|k] \\
    \mathbf{b}[k+1|k] &= \mathbf{b}[k|k] + \delta_k\dot{\mathbf{b}}[k|k]  + \frac{1}{2}\delta_k^2\ddot{\mathbf{b}}[k|k],\\    
\end{aligned}
\end{equation}
where $\delta_k$ is the time interval between time index $k+1$ and $k$. The update step and variance propagation and update are the same as the vanilla Kalman filter~\cite{barfoot2017state}. The aerodynamic drag coefficient $K_D$ is estimated via a Recursive Least-square estimator~\cite{muller2011quadrocopter} in an off-line fashion.

For the trajectory prediction, given the estimated ball state $\hat{\mathbf{s}}_b$ at time $t_0$, future ball trajectory predictions are made via numerically integrating \eqref{eqn:ball_motion} for a finite horizon. Hence, ball trajectory predictions will change when a new state measurement comes, and a new motion plan should be made. This requires that the motion planner can make new plans in a very short time (typically $20~ms$~\cite{lampariello2011trajectory}) so that the information of the latest ball trajectory prediction can be used in the latest robot motion plan. 
 
\subsection{Motion Planning}
In order to achieve better performance, the mobile platform trajectory needs to be updated when a new ball trajectory prediction is available. The motion planner is bi-level: \emph{1)} the \emph{high-level} planner calculates a feasible and optimal catching configuration $\mathbf{q}_f$ and catch time $t_f$ that enable the end effector to intercept the ball's trajectory at time $t_f$, and \emph{2)} the \emph{low-level} planner generates a smooth joint trajectory $\mathbf{q}_f(t)$ that takes the robot from its current to the desired catch configuration. Moreover, the motion planner must take into account kinematic redundancy, collision avoidance, and robot kinematic and dynamic constraints (e.g. position, velocity and torque constraints). This problem is a nonlinear and non-convex optimization problem~\cite{lampariello2011trajectory}. Solving such an optimization problem even off-line is still an open question in the field~\cite{bauml2010kinematically, lampariello2011trajectory}.

To tackle this planning problem in an on-line fashion, we need to make the following three simplifications. Similar simplifications are also made in the ball catching literature and work well in experiments~\cite{bauml2010kinematically, bauml2011catching, lampariello2011trajectory, Billard}.
\begin{simplification}[Arm-base collision] There are no nearby obstacles and no arm self-collision check; hence, no obstacle or self-collision check is needed. The only collision that may happen is the collision between the arm end effector and the mobile base. 
\end{simplification}

This simplification is reasonable because during robot catching experiments, usually the arm will extend to reach out for a catching. In such a scenario, arm self-collision is rare but collisions between the arm and mobile base are possible. Specifically, to prevent such a collision, we require the arm end effector to stay inside a semi-cylinder centered on the arm's base:
\begin{equation}
    \label{eqn:collision-sim}
    \begin{aligned}
        {_a}P_{ee,x}^2(\mathbf{q}) + {_a}P_{ee,y}^2(\mathbf{q}) \le R_{coll}^2 \\
        0 \le {_a}P_{ee,z}(\mathbf{q}) \le H_{coll}, 
    \end{aligned}
\end{equation}
where $ [{_a}P_{ee,x}, {_a}P_{ee,y}, {_a}P_{ee,z}] $ is the end effector's position in the arm frame which can be calculated via forward kinematics functions given the robot configuration $\mathbf{q}$, $R_{coll}$  is the cylinder's radius and $ H_{coll} $ is the cylinder's height. 

Through experiments, we noticed that as long as the end effector is inside this semi-cylinder space at catch time, the whole generated end effector trajectory usually will stay in that space. Thus constraints in \eqref{eqn:collision-sim} is only checked for $\mathbf{q}_f$ instead of every $\mathbf{q}(t)$ for $t_0 \leq t \leq t_f$.

\begin{simplification}[Kinematic planning]
We use kinematics models to depict the joint motion, specify objective functions and constraints on the joint accelerations, instead of torques. \end{simplification}

This simplification works in experiments as long as maximum joint acceleration values are chosen conservatively and low-level joint controls work properly. We use a double integrator as the kinematics model of a single joint and take joint acceleration as the control input:
\begin{equation}
\label{eqn:joint_motion_full_form}
        \begin{bmatrix}
  q_i[k+1]\\
  \dot{q}_i[k+1]
\end{bmatrix} = \begin{bmatrix}
  1 & \gamma\\
  0 & 1 
\end{bmatrix}  \begin{bmatrix}
  q_i[k]\\
  \dot{q}_i[k]
\end{bmatrix} + \begin{bmatrix}
  \frac{1}{2}\gamma^2\\
  \gamma
\end{bmatrix} u_{i}[k],
\end{equation}
where index $i$ represents the $i$-th joint, $\gamma$ is the step size\footnote{The step size $\gamma$ used for joint motion equations is not necessarily the same as the step size $\delta_k$ used for the Kalman filter}, and $u_{i}[k]$ is the control input. To ensure the trajectory feasibility, we have the following box constraints:

\begin{equation}
    \label{eqn:kinematics_constraints}
    \begin{aligned}
   \mathbf{q}_{min} \leq \mathbf{q}[k] \leq \mathbf{q}_{max}, \\
   \dot{\mathbf{q}}_{min} \leq \dot{\mathbf{q}}[k] \leq \dot{\mathbf{q}}_{max}, \\
   \ddot{\mathbf{q}}_{min} \leq \mathbf{u}[k] \leq \ddot{\mathbf{q}}_{max}, \\
    \end{aligned}
\end{equation}
where $ \mathbf{q}_{min}, \dot{\mathbf{q}}_{min}, \ddot{\mathbf{q}}_{min},  \mathbf{q}_{max}, \dot{\mathbf{q}}_{max}, \ddot{\mathbf{q}}_{max}$ are minimum and maximum value of joint positions, velocities and accelerations, respectively.

\begin{simplification}[Quadratic cost function] The cost functions in the high- and low-level planners are quadratic. We use quadratic cost functions because they are popular in the optimization field and usually easy to deal with.
\end{simplification}

With these three simplifications, we solve the motion planning problem in a bi-level fashion. The high-level planner selects the final catching configuration $\mathbf{q}_f$ and catch time $t_f$, while including the nonlinear non-convex constraints in~\eqref{eqn:probblem_formulation} and~\eqref{eqn:collision-sim}. Given the solution $\mathbf{q}_f^*, t_f^*$, the low-level path planner determines the intermediate trajectory waypoints $\mathbf{q}(t_k), t_0 < t_k < t_f$, but is restricted only by the linear constraints in \eqref{eqn:joint_motion_full_form} and \eqref{eqn:kinematics_constraints}. This is a quadratic program that can be solved very efficiently. To ensure that the optimal values $(\mathbf{q}_f^*, t_f^*)$ calculated by the high-level planner are feasible for the low-level planning problem, a box constraint on $\mathbf{q}_f$ which depends on $t_f$ and initial robot conditions is defined.

Compared to the parameterization methods used in literature for the low-level planning problem, like trapezoidal velocity profiles~\cite{bauml2010kinematically, bauml2011catching} and third-order polynomials~\cite{kocc2018online}, the QP technique we use allows more flexible trajectories due to the increased number of decision variables. 
Furthermore, via constraints of kinematic models in the quadratic program formulation, the generated trajectories are smoother and should be closer to the robot's actual behaviors when compared to trapezoidal velocity profiles and third-order polynomials. Details on the high- and low-level planners follow.

\subsubsection{High-level goal planner}
Given the initial robot configuration and velocity $(\mathbf{q}_0, \dot{\mathbf{q}}_0)$, ball position and velocity trajectory prediction $(\mathbf{b}(t), \dot{\mathbf{b}}(t))$ and initial time $t_0$, the high-level planner aims to find the optimal final catching configuration and catch time $(\mathbf{q}_f^*, t_f^*)$. Formally, this can be written as the following optimization problem: 
\begin{equation}
\label{eqn:high_level_opt}
\begin{aligned}
    (\mathbf{q}_f^{*}, t_f^*) = & \quad \argmin_{\mathbf{q}_f, t_f} \; \Gamma_h(\mathbf{q}_f, t_f) \\
    \text{s.t.} & \quad  {_a}P_{ee,x}^2(\mathbf{q}_f) + {_a}P_{ee,y}^2(\mathbf{q}_f) \le R_{coll}^2, \\
                & \quad  0 \le {_a}P_{ee,z}(\mathbf{q}_f) \le H_{coll}, \\
                & \quad _w\mathbf{P}_{ee}(\mathbf{q}_f) = \mathbf{b}(t_f) \\
                & \quad cos(<_w\mathbf{z}_{ee}(\mathbf{q}_f), - \frac{\dot{\mathbf{b}}(t_f)}{||\dot{\mathbf{b}}(t_f)||}>) = 1 \\
                & \quad   |\mathbf{q}_f - \mathbf{q}_0| \leq \Delta\mathbf{q}( \mathbf{q}_0, \dot{\mathbf{q}}_0, t_f - t_0),
\end{aligned}
\end{equation}
where $\Gamma_h$ is a user-defined quadratic cost function for the high-level planner, and $\Delta\mathbf{q}(\mathbf{q}_0, \dot{\mathbf{q}}_0, t_f - t_0)$ is the box constraint on $\mathbf{q}_f$ given initial joint position, velocity and duration time.

Accurately calculating $\Delta\mathbf{q}(\mathbf{q}_0, \dot{\mathbf{q}}_0, t_f - t_0)$ requires calculating the optimal intermediate waypoints $\mathbf{q}(t)$ between $\mathbf{q}(t_0)$ and $\mathbf{q}(t_f)$ under kinematic constraints \eqref{eqn:high_level_opt} for every catch time guess $t_f$ need to be calculated. This is too time-consuming and an approximation is needed. In this paper, we use the following approximation:
\begin{equation}
\label{eqn:appro_minmax_distance}
\begin{aligned}
     \Delta q_{i}(&q_{0, i}, \dot{q}_{0, i}, \Delta t)  =  \\
     &\lambda\left({ \dot{q}_{max, i} \left(\Delta t - \frac{\dot{q}_{max, i} - \dot{q}_{0, i}}{\ddot{q}_{max, i}}\right) + \frac{\dot{q}_{max, i}^2 - \dot{q}_{0, i}^2}{2\ddot{q}_{max, i}}}\right)\,,
\end{aligned}
\end{equation}
where $\Delta t = t_f - t_0$ is the duration time, $\dot{q}_{max, i}, \ddot{q}_{max, i}$ are the maximum velocity and acceleration of the $i$-th joint, $\lambda \in (0, 1)$ is a proportion parameter that needs to be tuned. The underlying assumption is that the maximum joint movement is proportional to the distance that the joint can traverse under the trapezoidal velocity profile~\cite{bauml2010kinematically}. Namely the joint first accelerates to its maximum speed and holds that speed until the end. We choose this approximation because of its simplicity and few corner cases needed to be considered in implementations. Other trajectory profiles like third-order polynomial can also be used for approximation~\cite{kocc2018online}. The SQP algorithm is used to solve the problem specified in~\eqref{eqn:high_level_opt}.

\subsubsection{Low-level path planner} \label{sec:low_level_planner} Given the desired configuration $\mathbf{q}_f^*$ and catch time $t_f^*$ from the high-level goal planner, the task of the low-level path planner is to generate a sequence of intermediate trajectory waypoints for each joint. 

For the $i$-th joint, this problem can be formulated as:
\begin{equation}
\label{eqn:low_level_opt}
\begin{aligned}
    \mathbf{U}_i^*  = &  \argmin_{\mathbf{U}_i} \; \Gamma_l(\mathbf{P}_i, \mathbf{U}_i) \\
    \text{s.t.} & \; \begin{bmatrix}
  q_i[k+1]\\
  \dot{q}_i[k+1]
\end{bmatrix} = \begin{bmatrix}
  1 & \gamma\\
  0 & 1 
\end{bmatrix}  \begin{bmatrix}
  q_i[k]\\
  \dot{q}_i[k]
\end{bmatrix} + \begin{bmatrix}
  \frac{1}{2}\gamma^2\\
  \gamma
\end{bmatrix} u_{i}[k] \\
    & \; q_{min,i} \leq q_i[k] \leq q_{max, i} \\
     & \; \dot{q}_{min, i} \leq \dot{q}_i[k] \leq \dot{q}_{max, i} \\
     & \; \ddot{q}_{min, i} \leq u_i[k] \leq \ddot{q}_{max, i} \\
    & \;q_i[0] = q_{0, i}, \dot{q}_i[0] = \dot{q}_{0, i} \\
    & \;q_i[K] = q_{f, i}^*,\dot{q}_i[K] = 0,
\end{aligned}
\end{equation}
where $\Gamma_l$ is a user-defined quadratic cost function for the low-level planner, $ K = floor( (t_f^* - t_0) / \gamma) $ is the finite horizon and $floor(x)$ returns the greatest integer less than or equal to $x$, $\mathbf{P}_i = [q_i[1], \cdots, q_i[K]]^T$ and $\mathbf{U}_i = [u_i[1], \cdots, u_i[K]]^T$  are the position and control input sequence of the $i$-th joint.

This is an optimal control problem that consists of quadratic cost functions and linear constraints. Standard quadratic programming methods can be used to solve this problem efficiently~\cite{boyd2004convex}. With the solution of optimal control input sequence $\mathbf{U}_i$ and initial conditions $q_i[0] = q_{0, i}, \dot{q}_i[0] = \dot{q}_{0, i}$, the optimal position sequence $\mathbf{P}_i$ can be constructed via the kinematic model in \eqref{eqn:joint_motion_full_form}. 

\subsection{Low-level Joint Control}
\label{sec:low-level-control}
\begin{figure}[tbph]
\centering
\includegraphics[width=1.0\linewidth]{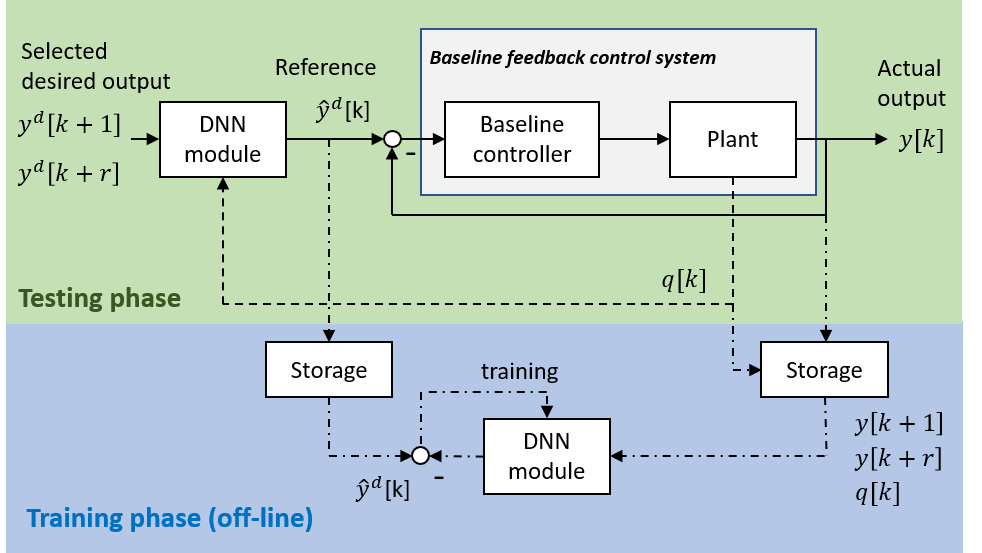}
\caption{A diagram of the inverse dynamics learning framework and the training phase. The DNN module modifies the reference signal to the baseline system for trajectory tracking performance improvement. Adapted from~\cite{zhou-cdc17}. }
\vspace{-10pt}
\label{fig:invlearning_diagram}
\end{figure}

Given the desired position trajectories generated by the motion planners, the task of the low-level joint controllers is to accurately track the trajectory. Accurate trajectory tracking is challenging due to the high-speed joint trajectories required for successful ball catching. Moreover, joint trajectory replanning may introduce discontinuities into the desired trajectories, which may be harder to track. To overcome these problems, we use the inverse dynamics learning technique developed in~\cite{zhou-cdc17}, shown in Fig.~\ref{fig:invlearning_diagram}. It uses DNNs as an add-on block to baseline feedback control systems to achieve a unity map between desired and actual outputs.  

This method learns an inverse dynamics model of the baseline closed-loop system: $\hat{y}[k] = f(x[k], x[k+r])$, where $x[k]$ is the system's discrete state at time index $k$, $\hat{y}[k]$ is the reference signal to the baseline control system, and $r$ is the relative degree~\cite{zhou-cdc17}. For discrete-time systems, the vector relative degree can be intuitively understood as the number of time steps before a change in the input is reflected in the output, which can be easily determined experimentally from the system’s step response~\cite{zhou-cdc17}.

Specifically, for a single robot joint of the mobile manipulator, the DNN's inputs and outputs are: 
\begin{equation}
    \hat{y}_j^d[k+1] = NN(q[k], y_j^d[k+r]) \,,
\end{equation}
where $y_j^d[k]$ is the reference signal sampled from desired trajectories for the $j$-th joint at time index $k$, and $\hat{y}_j^d[k+1]$ is the DNN's modified reference signal. As mentioned in~\cite{zhou-cdc17}, subtracting $y_j^d[k+1]$ from both the inputs and outputs of the neural network can improve the DNN's generalization ability since only the reference signal offsets $\tilde{y}_j^d[k+1] = \hat{y}_j^d[k+1] - y_j^d[k+1]$ instead of the reference signal itself $ \hat{y}_j^d[k+1] $ are learned:
\begin{equation}
\label{eqn:inv_dyn_mapping}
    \tilde{y}_j^d[k+1] = NN(q[k] - y_j^d[k+1], y_j^d[k+r] - y_j^d[k+1])\,.
\end{equation}
The  modified  position  reference  signal can then be  sent  to  the baseline control system to actuate the robot joint.

\section{Framework Implementation and Evaluation}
\label{sec:experiments}
This section presents implementation details and evaluation of each of the three framework components.

\subsection{Ball Estimation and Trajectory Prediction}
First, we describe the implementation details of the ball state estimation and trajectory prediction. The modified Kalman filter, stated in Section \ref{sec:ball_estimation_prediction}, is used to reduce noise in VICON position and velocity measurements. In experiments, VICON position measurements are very precise. Therefore, ball state estimation consists of position measurements from the VICON system and velocity estimation from the Kalman filter. Trajectory prediction is done through numerical integration with a step size of $0.1~ms$. 

Both the Kalman filter~\eqref{eqn:kalman_filter} and trajectory prediction \eqref{eqn:ball_motion} need an estimate of the aerodynamic drag coefficient $K_D$. To estimate it, we record ball state data from 20 random ball tosses and use a recursive least-squares estimator. The estimated value of $K_D$ is $0.0238~m^{-1}$. 

\subsection{Motion Planning}
\label{sec:results_motion_planning}
We first describe the implementation details of the high-level and low-level motion planner. The goal of the high-level planner is to find a catch configuration ${\bf{q}}_f$ and catch time $t_f$ while minimizing the robot movement:
\begin{equation}
    \label{eqn:cost-dist}  
    \Gamma_h(\mathbf{q}_f, t_f) = w_a \lVert \mathbf{q}_{a,0} - \mathbf{q}_{a, f} \rVert_2^2 + w_b \lVert \mathbf{q}_{b, 0} - \mathbf{q}_{b,f} \rVert_2^2\,, 
\end{equation}
where $w_a, w_b$ are weight parameters that penalize movement of the arm joints and the mobile base, respectively. As mentioned in Section \ref{sec:related_work}, control accuracy of the mobile base is usually lower than the arm \cite{kuka_base, ur10_arm}. Therefore, we select $w_a = 1, w_b = 5$ to reduce the mobile base motion. The proportion parameter used in the high-level planner~\eqref{eqn:appro_minmax_distance} is set to be $\lambda = 0.4$. The high-level planner uses the SQP algorithm which we implemented using the NLopt c++ library~\cite{Nlopt}. The initial guess for the optimization variables $(\mathbf{q}_f, t_f)$ are the robot's current configuration and a hard-coded catch time guess $t_f = 0.5~s$. 

The goal of the low-level planner is to generate a smooth trajectory from current configuration ${\bf{q}}_0$ to catch configuration ${\bf{q}}_f$. This is achieved by penalizing joint acceleration change from one time step to the next during the trajectory, which can be written as follows for the $i$-th joint:
\begin{equation}
    \Gamma_l(\mathbf{P}_i, \mathbf{U}_i) = \sum_{k=0}^{K-1}||u_i[k+1] - u_i[k]||_2^2\,,
\end{equation}
where $K$ is calculated according to Section~\ref{sec:low_level_planner}, and the discretization step size is $\gamma = 0.05~ms$. We solve the QP in the low-level planner with qpOASES c++ library~\cite{Ferreau2014}.


We use numerical experiments to evaluate the motion planner's performance since we cannot control the ball initial state in real experiments when the ball is thrown by a human operator. We generated $2560$ initial ball state samples, whose positions are uniformly distributed in a circle around the robot. In order to ensure that the resulting ball trajectories are feasible given the robot's motion constraints, we first note that the end effector can move $0.65~m$ in $0.86~s$, as described in Section~\ref{sec:ball_catching_exp}. We restrict the initial ball states to be such that after $0.7~s$ the ball is predicted, via numerical integration of \eqref{eqn:ball_motion}, to be located inside a $0.25~m$ diameter sphere centered around the end effector of the given robot configuration. 

The proposed motion planner can find a trajectory for $95.23\%$ of all initial ball samples. The average computation time is $5.10~ms$, with $4.80~ms$ used for the high-level nonlinear planner, and $0.3~ms$ used for the low-level (quadratic program) planner on an Intel Xeon CPU i7-8850H 2.60GHz computer. No sophisticated techniques like multi-threaded programming are used to speed up computation time.

\subsection{Low-level Control}
We first describe the implementation of the learning-based low-level controller. We first estimate experimentally the relative degree $r$ of each joint, which is $8$ for arm joints and $14$ for mobile base joints.

To learn the mapping function in \eqref{eqn:inv_dyn_mapping}, we train $8$ DNNs, one for each joint. Each DNN consists of two hidden layers and each layer contains $10$ neurons. The ReLU function is used as the activation function.  To create the training and testing dataset, we use the low-level path planner (described in Section~\ref{sec:low_level_planner}) to generate trajectories similar to those used in real ball catching experiments. We characterize these trajectories by bounding the maximum joint displacement $\Delta \mathbf{q}$ in a given amount of time $\Delta t$. In our experiments we set  $\Delta t = {2s, 3s, 4s}$, to represent trajectories with different speeds and $\Delta \mathbf{q}=[1.36, 1.36, 1.36, 1.75, 1.75, 2.53, 0.76, 0.76]^T$, where the first six  joint displacements correspond to the arm and are specified in radians, and the last two joint displacements correspond to the mobile base and are specified in meters. 

The trajectories are executed using the robot's baseline control system. It consists of off-board PD controllers that generate desired joint velocities and on-board controllers of the UR10 arm and mobile base that turn desired velocities into joint torques (UR10 arm) or wheel rotation velocities (mobile base). In this way we collect a dataset with $1000$ data pairs for each joint. The dataset is divided with $80\%$ is used for training and the rest is used for validation. We use Adam optimizer implemented in \textit{Tensorflow} to train the neural network with $50$ training iterations. 

To test the influence of inverse dynamics learning on tracking performance, we generate testing trajectories by setting $\Delta \mathbf{q}=[0.8, 0.8, 0.8, 1.1, 1.1, 1.1, 0.20, 0.20]^T$ and $\Delta t = 1~s$, which are different from training and validation trajectories in the dataset. The testing trajectories are executed ten times each when the robot uses the learning technique and ten times each when no learning technique is used. The trajectory tracking rooted-mean-squared error (RMSE) over the ten trajectories is shown in Table \ref{tab:joint_tracking_error}. It can be seen that the inverse dynamics learning method can reduce tracking errors between $35\% \sim 80\%$ with the highest error reduction corresponding to the mobile base.

\begin{table}[!htb]
\centering
\captionsetup{justification=centering}
\caption{Joint trajectory tracking error comparison}
\label{tab:joint_tracking_error}
\begin{tabular}{rccc}
\toprule
\rowcolor{Gray}
Joint &  With DNN [$^\circ$]& Without DNN [$^\circ$] & Reduction\\
\midrule
arm 1 &  $\mathbf{0.0218}$ &  $0.0508$ & $57.2\%$ \\
2 &  $\mathbf{0.0300}$ &  $0.0508$ &  $40.9\%$ \\
3 &  $\mathbf{0.0264}$ &  $ 0.0506$ &  $47.8\%$ \\
4 &  $\mathbf{0.0309}$ &  $0.0561$ &  $45.0\%$ \\
5 &  $\mathbf{0.035}$ &  $0.0562$ &  $36.5\%$ \\
6 &  $\mathbf{0.034}$ &  $0.0561$ & $37.1\%$ \\
\toprule
\rowcolor{Gray}
Joint &  With DNN [$m$]& Without DNN [$m$] & Reduction\\
\midrule
base $x$ & $\mathbf{0.0041}$ & $0.0196$ & $79.1\%$ \\
$y$ & $\mathbf{0.0047}$ & $0.0183$ & $74.0\%$ \\
\bottomrule
\end{tabular}
\end{table}

\section{Experiments}
This section presents the experimental setup and ball catching experimental results. A video demonstrating the real robot experiments can be found at \url{http://tiny.cc/ball_catch}.

\subsection{Experimental Setup}
\label{sec:general_setup}
The mobile manipulator used in this experiment consists of a 6-DOF UR10 arm and a 3-DOF Ridgeback omnidirectional mobile base driven by four Mecanum wheels. We use symmetric joint motion boundary values, i.e. $\mathbf{q}_{min} = -\mathbf{q}_{max},\ \dot{\mathbf{q}}_{min} = -\dot{\mathbf{q}}_{max},\ \ddot{\mathbf{q}}_{min} = -\ddot{\mathbf{q}}_{max}$. The six arm joints have the same maximum joint positions of $180^\circ$ and accelerations of $458{^\circ/s^2}$. Their maximum joint velocities are $103{^\circ/s}$, $103{^\circ/s}$, $103{^\circ/s}$, $126{^\circ/s}$, $126{^\circ/s}$, and $180{^\circ/s}$ for arm joint $1\sim6$ respectively. The maximum positions, velocities and accelerations of the mobile base in the $x$ and $y$ axes are $3m$, $1.0m/s$, $2.5m/s^2$, respectively.

\begin{figure}[tbph]
\centering
\includegraphics[width=0.8\linewidth]{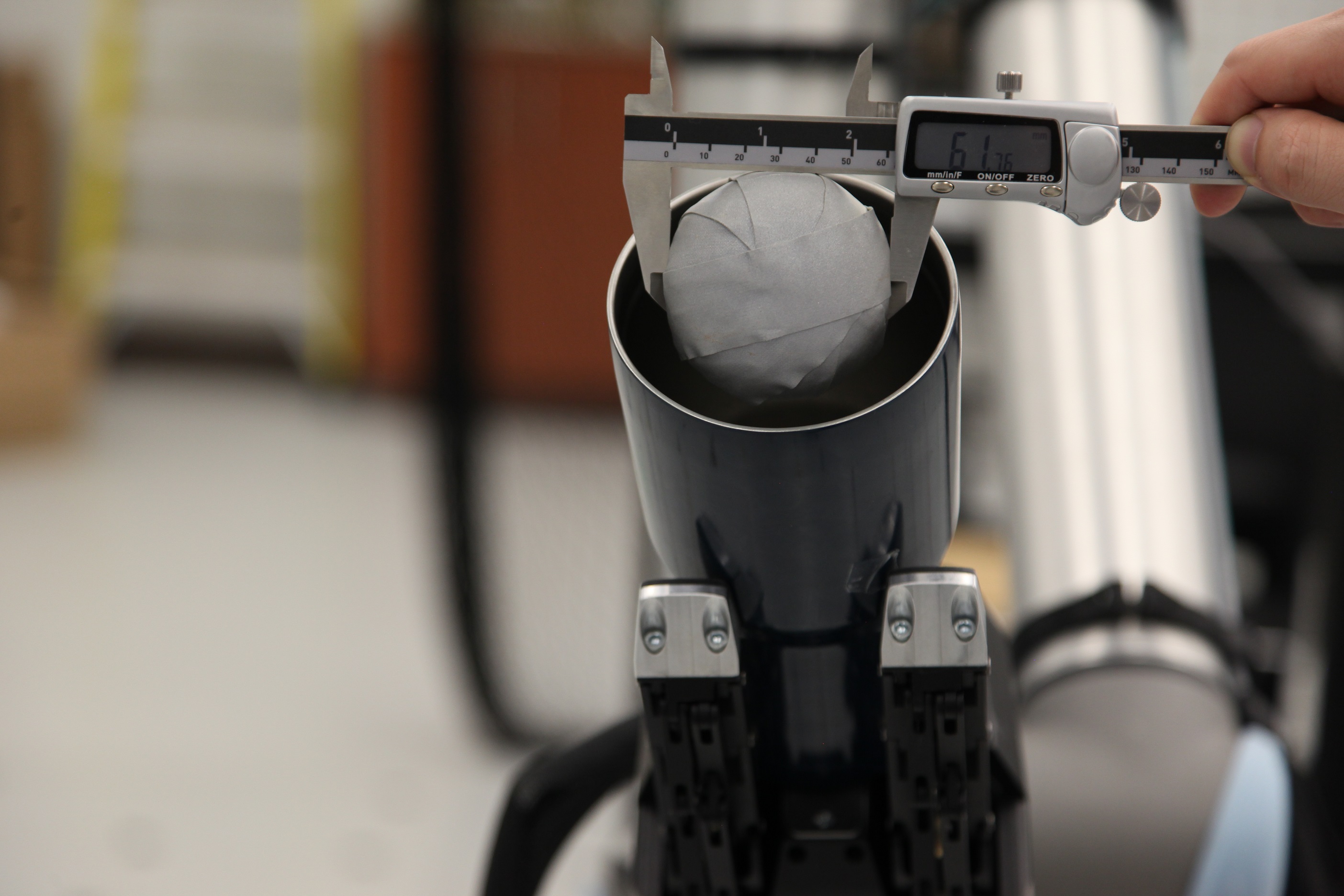}
\caption{The $61.76~mm$ diameter ball and $94.90~mm$ diameter cup used in this paper. The position accuracy requirement in the $x,y,z$ axes is approximately $(94.90 - 61.76)/2=16.57~mm$. The typical speed at the catch time is around $5~m/s$, leading to the $16.57/5~\approx~3~ms$ time precision requirement.}
\vspace{-10pt}
\label{fig:ball_cup}
\end{figure}

The three-finger end effector grasps a $94.90~mm$ diameter metal cup that is used to catch the ball, as shown in Fig.~\ref{fig:ball_cup}.  The proposed framework is implemented via Robot Operation System (ROS) and runs on a ThinkPad P52 laptop. The ball used in these experiments is a standard $61.76~mm$ tennis ball with a weight of $ 81.76~g$. The ball is wrapped with retro-reflective tape in order to be visible to the VICON motion capture system, which provides position and velocity measurements to be used in the ball state estimation module.

\subsection{Ball Catching Experiment}
\label{sec:ball_catching_exp}
We conducted experiments to test three main characteristics of the proposed framework: \emph{(i)} ball catching success rate, \emph{(ii)} impact of the learning-based controller, and \emph{(iii)} impact of QP-based low-level planner. 
\subsubsection{Experiment A: Ball Catching Success Rate} 
\begin{figure}[tbph]
\centering
\includegraphics[width=1.0\linewidth]{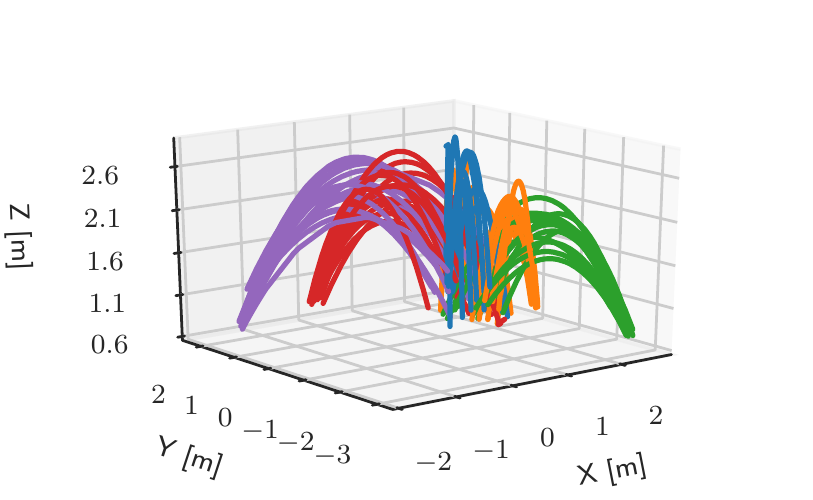}
\caption{Ball trajectories of Experiment A. Starting from the left corner and in the anti-clockwise direction, the five ball throwing positions in the $xy$ plane in meters are $(2.8, -3.1), (2.9, 0.0), (2.5, 2.0), (0.0, 2.8), (-2.0, 2.0)$. Ball trajectories are colored according to its throwing location.}
\vspace{-10pt}
\label{fig:ball_trajectory}
\end{figure}

\begin{figure}[htp]
	\centering
	\begin{subfigure}{0.5\textwidth}
	    \captionsetup{justification=centering}
		\includegraphics[width=0.8\textwidth]{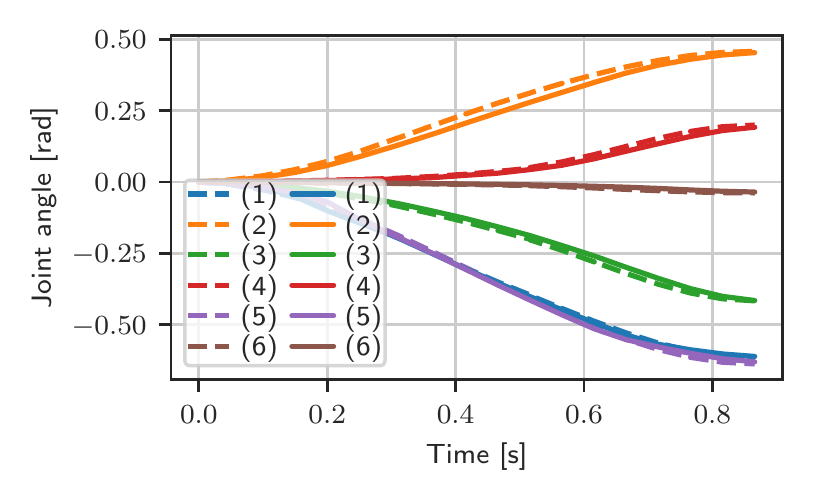}%
		\centering
		\caption{}
		\label{fig:arm_tracking}
	\end{subfigure}\\
	\begin{subfigure}{0.5\textwidth}
	\captionsetup{justification=centering}
	    \centering
		\includegraphics[width=0.8\textwidth]{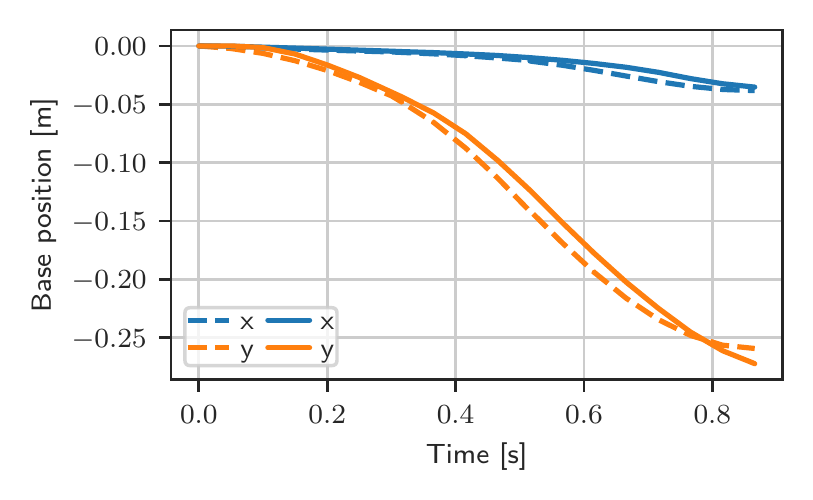}%
		\caption{}
		\label{fig:base_tracking}
	\end{subfigure}
	\caption{An example of \emph{(a)} arm joint and \emph{(b)} mobile base tracking performance. Dashed lines represent desired trajectories, while solid lines represent actual trajectories. For a better illustration effect, we subtract initial values from trajectory waypoints, and thus joint angles relative to their initial values are shown here.}
	\vspace{-10pt}
	\label{fig:tracking_performance}
\end{figure}

In this experiment, we use the framework proposed in Section \ref{sec:methodology} and throw balls from different locations while the robot has different initial configurations. Three initial robot configurations are chosen. For each initial robot configuration, we throw the ball from five different locations, five throws for each location, for a total of 75 ball catching experiments. All ball trajectories are shown in Fig.~\ref{fig:ball_trajectory}. Of the $75$ ball throws, $65$ balls are caught successfully by the robot, which is a $85.33\%$ success rate. The end effector position trajectory tracking RMSE over the $75$ experiments are $11.03~mm, 14.01~mm, 13.25~mm$ for $x, y, z$ axes respectively. Among the successful catches, the maximum end effector displacement in a catch is $0.65~m$ in $0.86~s$ with a mobile base displacement of $0.26~m$ in that catch. The corresponding arm and base position trajectories for this catch are shown in Fig.~\ref{fig:arm_tracking} and \ref{fig:base_tracking}. It can be seen that our mobile base is able to move accurately in both $x$ and $y$ directions while the humanoid in \cite{bauml2011catching} can only move in one direction.

\subsubsection{Experiment B: Impact of Learning-based Controller} In this experiment, we are interested in showing that the learning-based controller improves tracking performance, therefore improving the success rate of the overall framework. It is hard to repeat a ball trajectory twice because a human throws it. To do a fair comparison in this experiment we throw balls from the same location $(2.5, 2.0, 0)$ while the robot has the same initial configuration, as shown in Fig.~\ref{fig:geo-setup}. We make $40$ throws when the robot uses inverse dynamics learning and $40$ when the robot does not use inverse dynamics learning for a total of $80$ throws. The success rate for this particular configuration is $75\%$ when using inverse dynamics learning and $65\%$ when not using it. The ten percent decrease shows the effectiveness of inverse dynamics learning. Note that the success rate for this particular configuration is lower than the success rate in Experiment A. Visual inspection of the data and video show that the geometrical setup, i.e. ball throwing location and robot initial configuration, used in this experiment requires more robot maneuvers than other geometrical setups used in Experiment A.

\subsubsection{Experiment C: Impact of QP-based Low-level Planner}
In this experiment, we are interested in showing the impact of the QP-based low-level planner in the ball catching success rate. To do this, we implemented the trapezoidal velocity profile method in~\cite{bauml2010kinematically} for the low-level planner, and tested it by throwing $40$ balls with the geometrical setup as in Experiment B. Note that the inverse dynamics learning is still used. The robot achieves a success rate of $72.5\%$, similar to the success rate of $75\%$ when using the QP method.

The QP method does not have a significant impact on ball catching success rate. One possible reason is that the way the robot accelerates and decelerates does not have a large impact on the ball catching success due to the short duration of the trajectory~\cite{bauml2010kinematically}. However, the QP method does allow for more flexible and smooth trajectories\rev, which may improve the performance of tasks with longer duration. Finally, similar success rates with the two different low-level planners may be due to the high accuracy of the learning-based controller when tracking arbitrary trajectories.

\subsection{Failure analysis}
The success of ball catching tasks requires a tight integration of every system component. The malfunction of any component will result in a failure. Among all throws in Experiment A, B, C, failure has occurred for the following reasons: inaccurate ball trajectory prediction, large tracking errors, and failure to find feasible solutions of the high-level planning problem. Large tracking errors occur when maneuvers are too aggressive and overshooting happens. We could improve this by penalizing aggressive trajectories.

\section{Conclusion}
\label{sec:conclusion}
We presented a framework for accurate high-speed motion generation and execution on mobile manipulators with the application scenario of ball catching.  A modified Kalman filter that considers aerodynamics is used for estimating the ball's velocity, and ball trajectory prediction is done via numerical integration. A bi-level optimization scheme is proposed to handle complex trajectory requirements in real-time, with an average run time of $5.1~ms$. The high-level goal planning problem is formulated as a non-linear optimization problem and solved by SQP. The low-level path planning problem is solved by QP for more flexible and smooth trajectories. The joint control is driven by an inverse dynamics learning method. This learning-based controller reduces joint tracking errors by up to $79.1\%$, which translate into a $10\%$ increase in success rate compared to a non-learning-based joint controller. The proposed framework is validated via extensive real robot catching experiments under different configurations and achieves a success rate of $85.33\%$, setting a new state of the art. In future work we aim to focus on dynamic planning instead of kinematic planning, which should improve the feasibility of joint trajecotries.

\bibliographystyle{IEEEtran}
\bibliography{reference}

\end{document}